\documentclass[11pt,a4paper]{article}
\usepackage[hyperref]{naaclhlt2019}
\usepackage{times}
\usepackage{latexsym}
\usepackage{graphicx}
\usepackage{amsmath}
\usepackage{subcaption}
\usepackage{upgreek}
\usepackage{pbox}
\usepackage{enumitem}

\usepackage{url}

\aclfinalcopy 
\def\aclpaperid{1456} 

\title{A Crowdsourced Frame Disambiguation Corpus with Ambiguity}

\author{Anca Dumitrache \\
  FD Mediagroep, \\
  Vrije Universiteit Amsterdam \\
  Netherlands \\
  {\tt anca.dmtrch@gmail.com} \\\And
  Lora Aroyo \\
  Google \\
  USA \\
  {\tt l.m.aroyo@gmail.com} \\\And
  Chris Welty \\
  Google \\
  USA \\
  {\tt cawelty@gmail.com} \\}

\date{}

\begin{document}
\maketitle
\begin{abstract}
  We present a resource for the task of FrameNet semantic frame disambiguation of over 5,000 word-sentence pairs from the Wikipedia corpus. The annotations were collected using a novel crowdsourcing approach with multiple workers per sentence to capture \textit{inter-annotator disagreement}. In contrast to the typical approach of attributing the best single frame to each word, we provide a list of frames with disagreement-based scores that express the confidence with which each frame applies to the word.  This is based on the idea that inter-annotator disagreement is at least partly caused by ambiguity that is inherent to the text and frames. We have found many examples where the semantics of individual frames overlap sufficiently to make them acceptable alternatives for interpreting a sentence. We have argued that ignoring this ambiguity creates an overly arbitrary target for training and evaluating natural language processing systems - if humans cannot agree, why would we expect the correct answer from a machine to be any different? To process this data we also utilized an expanded lemma-set provided by the Framester system, which merges FN with WordNet to enhance coverage.  Our dataset includes annotations of 1,000 sentence-word pairs whose lemmas are not part of FN. Finally we present metrics for evaluating frame disambiguation systems that account for ambiguity.
\end{abstract}

\section{Introduction}

Crowdsourcing has been a popular method to collect corpora for a variety of natural language processing tasks~\cite{Snow:2008}, although one of its downsides is the crowd's lack of domain knowledge that is helpful in solving some tasks. \textit{Semantic frame disambiguation} is an example of a complex natural language processing task that is usually performed by linguistic experts, subjected to strict annotation guidelines and quality control~\cite{baker2012framenet}. The theory of frame semantics~\cite{Fillmore:1982} defines a \textit{frame} as an abstract representation of a word sense, describing a type of entity, relation, or event, together with the associated \emph{roles} implied by the frame. The FrameNet (FN) corpus~\cite{baker1998berkeley} is a collection of semantic frames, together with a corpus of documents annotated with these frames. Similarly to word-sense disambiguation, \textit{frame disambiguation} is the task of obtaining the correct frame for each word, since many words have multiple possible meanings.

Using domain experts for frame disambiguation is expensive and time consuming, resulting in small corpora for this task that do not scale well for modern machine learning methods -- FN version 1.7, the latest one at the time of writing, contains only about 10,000 sentences annotated with frames. Furthermore, only using one expert to perform the annotation makes it difficult to capture any diversity of perspectives.  

There have been a number of small-scale attempts at using crowdsourcing for frame disambiguation in sentences, showing that the crowd has comparable performance to the FN domain experts~\cite{Hong:2011:GCR:2018966.2018970}, and that the crowd can be used to correct wrong examples that have been collected automatically~\cite{pavlick2015framenet+}. Crowd performance can be improved by combining frame role identification with disambiguation ~\cite{fossati2013outsourcing}, or by asking crowd workers to give each other feedback and then letting them change their answer~\cite{chang2015scaling}. Crowdsourcing has also been useful to identify the ambiguity in frame disambiguation~\cite{jurgens2013embracing}.  

Previously, we have shown~\cite{DBLP:conf/hcomp/DumitracheAW18} that while the crowd and FN expert mostly agree over frame disambiguation, disagreement cases are often caused by ambiguity, such as vague or overlapping frame definitions, or incomplete information in the sentence. Because of these issues with the input data, the approach of selecting one single correct frame for every word, and ignoring alternative interpretations, often results in arbitrary, incomplete ground truth corpora. In order to aggregate annotated data while preserving disagreement, we use the CrowdTruth method\footnote{\url{http://crowdtruth.org}}~\cite{aroyo2014threesides}, which encourages using multiple crowd annotators to perform the same work, and processes the disagreement between them to signal low quality workers, sentences, and frames.

This paper presents a crowdsourced FN frame disambiguation corpus of 5,042 sentence-word pairs (which has since grown to over 9,000 since the submission of this paper). More than 1,000 of these are lexical units (LUs) not part of FN. To our knowledge, it is the largest corpus of this type outside of FN. In addition, we applied the CrowdTruth method, in which each sentence and lexical item is accompanied by \emph{a list of multiple frames} with scores that express the confidence with which each frame applies to the word. This allows us to demonstrate that ambiguity is a prominent feature of frame disambiguation, with many cases where more than one possible frame can apply to the same word. Finally, we present an evaluation of several frame disambiguation models using evaluation metrics that leverage the multiple answers and their confidence scores, and show that even a model that always predicts the top crowd answer will not always have the best performance.

\section{Corpus Collection \& Analysis}
\label{sec:corpus}

\subsection{Data Preprocessing}

Our corpus consists of 5,042 candidate word-sentence pairs from Wikipedia (which has since grown to over 9,000 since the submission of this paper) and a candidate list of frames for the word, with 742 unique frames and 1,705 unique lexical units (LUs). The sentences have been randomly selected, based on these criteria:

\begin{itemize}
    \item The candidate word has \textit{no more than 25 candidate frames}, to not overwhelm the annotators.
    \item The part of speech of the word is a \textit{verb}.
    \item The \textit{distribution of candidate frames was optimized for maximum diversity} using a greedy approach.
\end{itemize}

To gather the candidate frames for each word, we gathered the candidate frames associated with the LU from FN1.7. Next we completed the candidate list using Framester~\cite{gangemi2016framester}, which maps FN semantic frames to synonym sets from WordNet~\cite{miller1995wordnet}. The sentences were processed with tokenization, sentence splitting, lemmatization and part-of-speech tagging. Then each word with a frame attached to it was matched with all of its possible synonym sets from WordNet, while making sure that the part-of-speech constraint of the synonym set is fulfilled. Using the WordNet mapping, we constructed the list of additional candidate frames for each word.  Framester disambiguation used release 1.5 of FN, and some frames changed names in version 1.7, so we manually mapped these frames from FS to their latest version. Framester disambiguation was also used to collect a subset of our corpus consisting of 1,000 sentence-word pairs with LUs that are not part of the FN corpus. For simplicity, we refer to the sentence-word pairs as sentences in the rest of the paper.

\subsection{Crowdsourcing Setup}

We ran the task on Amazon Mechanical Turk, where the workers were asked to select \emph{all frames} that fit the sense of the highlighted word in a sentence from the multiple choice candidate list, or that none of the frames is correct. We used 15 workers/sentence that were paid \$0.05 for each judgment, and a total cost of \$1.35 per sentence (after factoring in the additional AMT costs).\footnote{\url{https://mturk.com/}}

\begin{table*}[tbh!]
    \centering
    \scalebox{0.75}{
    \begin{tabular}{cp{10cm}cp{8cm}}
        \hline
        \# & \textsc{Sentence} & \textbf{$SQS$} & \textsc{Frames ($FSS$)}  \\ \hline
        1 & Domestication of plants has, over the centuries \textbf{improved} disease resistance. & 0.652 & \textit{improvement or decline} (0.823), \newline \textit{cause to make progress} (0.683) \\
        2 & He is the 5th of 8 male players in history to \textbf{achieve} this. & 0.626 & \textit{accomplishment} (0.764), \newline \textit{successful action} (0.709) \\
        3 & Albertus Magnus, a Dominican monk, \textbf{commented} on the operations and theories of alchemical authorities. & 0.511 & \textit{communication} (0.522), \newline \textit{statement} (0.703) \\
        4 & He \textbf{slices} at Hector's armor, throwing him off guard and spinning him around. & 0.319 & \textit{part piece} (0.499), \textit{cause harm} (0.4), \textit{cutting} (0.394), \textit{attack} (0.254), \textit{hit target} (0.227) \\
        5 & Another 46 steps \textbf{remain} to climb in order to reach the top, the ``terrasse'', from where one can enjoy a panoramic view of Paris. & 0.308 & \textit{left to do} (0.497), \textit{remainder} (0.478), \textit{state continue} (0.319), \textit{existence} (0.155) \\
        6 & Borzoi males frequently \textbf{weigh} more. & 0.283 & \textit{assessing} (0.421), \textit{dimension} (0.402), \newline \textit{importance} (0.128) \\ 
        7 & The dance includes bending and \textbf{straightening} of the knee giving it a touch of Cuban motion. & 0.24 & \textit{reshaping} (0.495), \textit{arranging} (0.356), \textit{body movement} (0.298), \textit{cause motion} (0.249) \\ 
        \hline
    \end{tabular}
    }
    \caption{Example sentences with disagreement over the frame annotations (candidate word in bold).}
    \label{tab:examples}
\end{table*}


To aggregate the results of the crowd while also capturing inter-annotator disagreement, we use the CrowdTruth metrics\footnote{\url{https://github.com/CrowdTruth/CrowdTruth-core}}~\cite{dumitrache2018crowdtruth}, replicating the setup from our previous work~\cite{DBLP:conf/hcomp/DumitracheAW18}. The choice of frames of one worker over one sentence are aggregated into a \textit{worker vector} -- a binary vector with $n+1$ components, where $n$ is the number of frames shown together with the sentence, where the decision to pick each of the frames (or none) corresponds to a component in the vector. The vectors are used to calculate quality scores for workers, sentences and frames. Although we make all quality scores available as part of the corpus, in this paper we focus on:


\begin{itemize}

\item \textbf{frame-sentence score ($FSS$):} the degree with which a frame matches the sense of the word in the sentence. It is the ratio of workers that picked the frame to all the workers that read the sentence, weighted by the worker quality.  A high $FSS$ means the frame is clearly expressed in a sentence.

\item \textbf{sentence quality ($SQS$):} the overall worker agreement over one sentence. It is the average cosine similarity over all worker vectors for one sentence, weighted by the worker quality and frame quality. A high $SQS$ indicates a clear sentence.

\end{itemize}

The aggregated crowdsourcing results and the FN 1.5 to 1.7 mapping table are available online.\footnote{\url{https://github.com/CrowdTruth/FrameDisambiguation}}

\subsection{Ambiguity in the Corpus}

\begin{figure}[t!]
\begin{subfigure}{.24\textwidth}
\centering
\includegraphics[width=\linewidth]{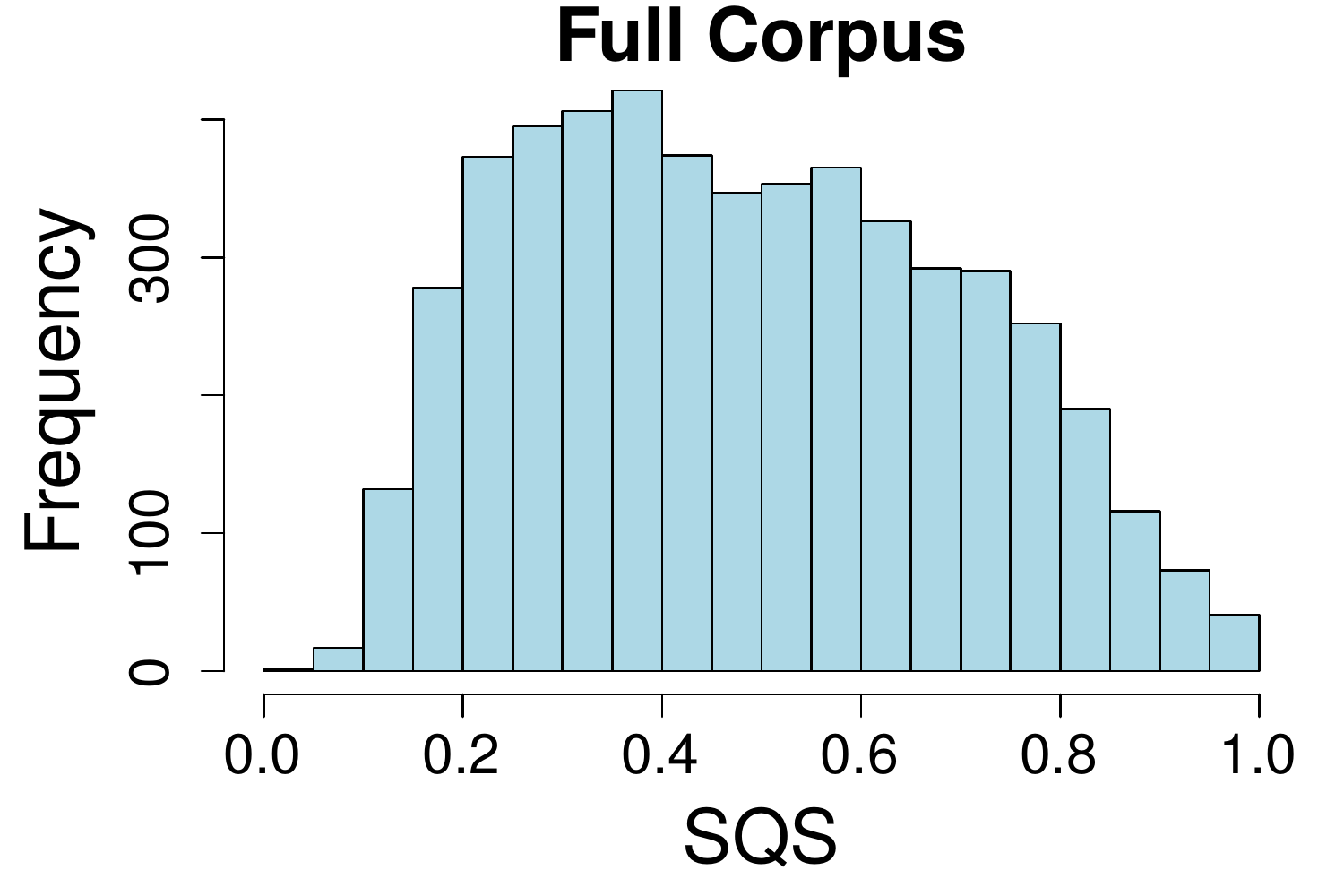}
\end{subfigure}%
\begin{subfigure}{.24\textwidth}
\includegraphics[width=\linewidth]{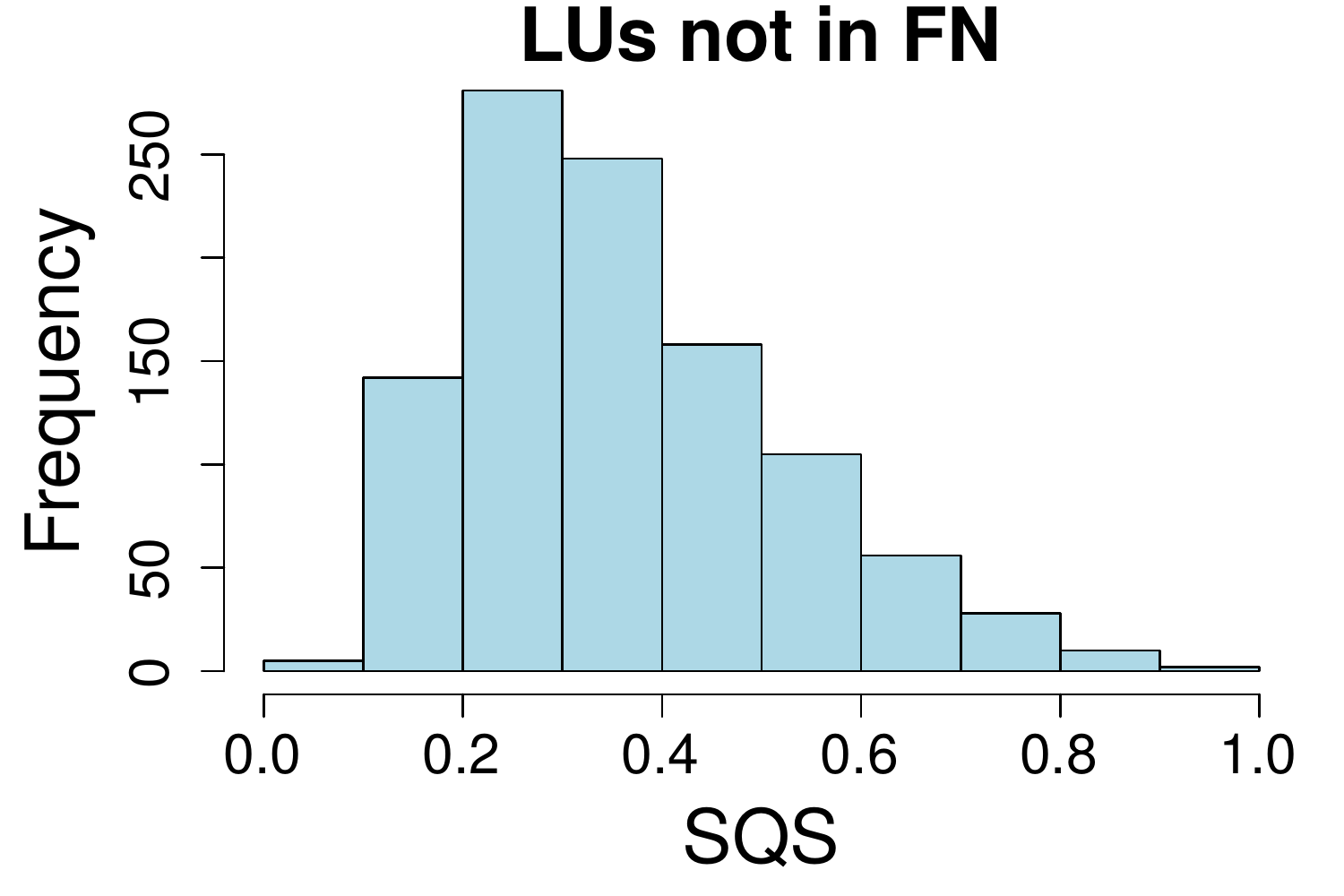}
\end{subfigure}
\caption{Histogram of $SQS$ values - the quality scores in sentences where the LU is not in FN skew lower.}
\label{fig:sqs_histo}
\end{figure}

An analysis of the corpus found many examples of inter-annotator disagreement, of which a few examples are shown in Table~\ref{tab:examples}. For 720 sentences, a majority of the workers picked at least 2 frames (examples 1-3 in Table~\ref{tab:examples}). And for 1,514 sentences, no one frame has been picked by a majority of the workers (examples 4-7 in Table~\ref{tab:examples}). Disagreement is also more prominent in the sentences where the LU is not a part of FN (Figure~\ref{fig:sqs_histo}).

The disagreement comes from a variety of causes: a parent-child relation between the frames (\textit{statement} and \textit{communication} in \#3), an overlap in the definition of the frames (\textit{accomplishment} and \textit{successful action} in \#2), the meaning of the word is expressed by a composition of frames (in \#7, ``straightening of the knee'' is a combination of \textit{reshaping} the form of the knee, \textit{arranging} the knee in the right position, and \textit{body movement}), and combinations of all of these reasons (in \#4, ``slices'' is a combination of \textit{part piece} and \textit{cause harm}, and the other frames are their children). More example sentences for each type of disagreement are available in the appendix. The sentences themselves are not difficult to understand, and it can be argued that all of them have one frame that applies the best for the word. The goal of this corpus is to show that next to this best frame for the word, there are other frames that apply to a lesser degree, or capture a different part of the meaning. When evaluating a model for frame disambiguation, it seems unfair to penalize misclassifications of frames that still apply to the word, but with less clarity, in the same way we would penalize a frame that captures a wrong meaning. Also, we argue that models should take into account that annotators do not agree over some examples, and treat them differently than clear expressions of frames. Disagreement can also be caused by worker mistakes (in \#6, \textit{dimension} refers to the size of the object, not the act of measuring the size). While we try to mitigate for this by weighing confidence scores with the worker quality, the mistakes still appear in the corpus. This type of disagreement could be useful in future work to identify examples that workers need to be trained on.



\section{Evaluating Frame Disambiguation}

\subsection{Systems Tested}

As an example usage of our corpus, we used it to evaluate these frame disambiguation models:

\begin{enumerate}
    \item \textbf{OS:} The Open-Sesame~\cite{swayamdipta:17} classifier, pre-trained on the FN corpus (release 1.7). Given a word-sentence pair, OS uses a BiLSTM model with a softmax final layer to predict a single frame for the word. If the LU is not in FN, it cannot make a prediction. 
    
    \item \textbf{OS+:} We modified the OS classifier to perform multi-label classification. To calculate the confidence score for candidate frame $f$, we removed the softmax layer and passed the output of the BiLSTM model $\nu(f)$ through the following transformation: $c(f) = [1 + tanh \ \nu(f)] / 2$. This gave a score $c(f) \in [0,1]$ expressing the confidence that frame $f$ is expressed in the sentence.
    
    \item \textbf{FS:} Framester includes a  tool for rule-based multi-class multi-label frame disambiguation~\cite{gangemi2016framester}. While for the dataset pre-processing (Sec.~\ref{sec:corpus}) we considered the frames for all synsets a word is part of, FS performs an additional word-sense disambiguation step 
    to return a more precise list of frames. We used the tool with \textit{profile T} as it was shown to have the overall better performance. FS can only predict FN frames from the 1.5 release, which is missing 202 frames from version 1.7. 
\end{enumerate}

While OS+ produces confidence scores, the other methods produce binary labels for each frame-sentence pair. These models do not have state-of-the-art performance~\cite{hermann2014semantic,fitzgerald2015semantic}, we picked them because they were accessible and allowed testing on a novel corpus. Finally, we evaluate the quality of the \textbf{TC} corpus, containing only the top frame picked by the crowd for every sentence. This test shows what is the best possible performance over our corpus that can be expected from a system such as OS that selects a single frame per sentence. 

\subsection{Evaluation Metrics \& Results}

Instead of traditional evaluation metrics that require binary labels, we propose an evaluation methodology that is able to consider multiple candidate frames for each sentence and their quality scores. We use \textbf{Kendall's $\uptau$} list ranking coefficient~\cite{kendall1938new} and \textbf{cosine similarity} to calculate the distance between the list of frames produced by the crowd labeled with the $FSS$, and the frames predicted by the baselines in each sentence. Whereas Kendall's $\uptau$ only accounts for the ranking of the $FSS$ for each frame, cosine similarity uses the actual $FSS$ values in the calculation of the similarity. Both metrics compute a score per sentence (Kendall's $\uptau \in [-1,1]$, and cosine similarity $\in [0,1]$).  Using these metrics, we produce two aggregate statistics over our test corpus: (1) the area-under-curve ($AUC$) for each metric, normalized by the corpus size, and (2) the $SQS$-weighted average of each metric ($w-avg$), which also accounts for the ambiguity of the sentence as expressed by the $SQS$. We evaluate on two versions of the corpus: (1) the restricted set (\textsc{R-Set}) of 4,000 sentences with LUs from the FN corpus, and (2) the full set (\textsc{F-Set}) of 5,042 sentences.

\begin{figure}[t!]
\begin{subfigure}{.24\textwidth}
\centering
\includegraphics[width=\linewidth]{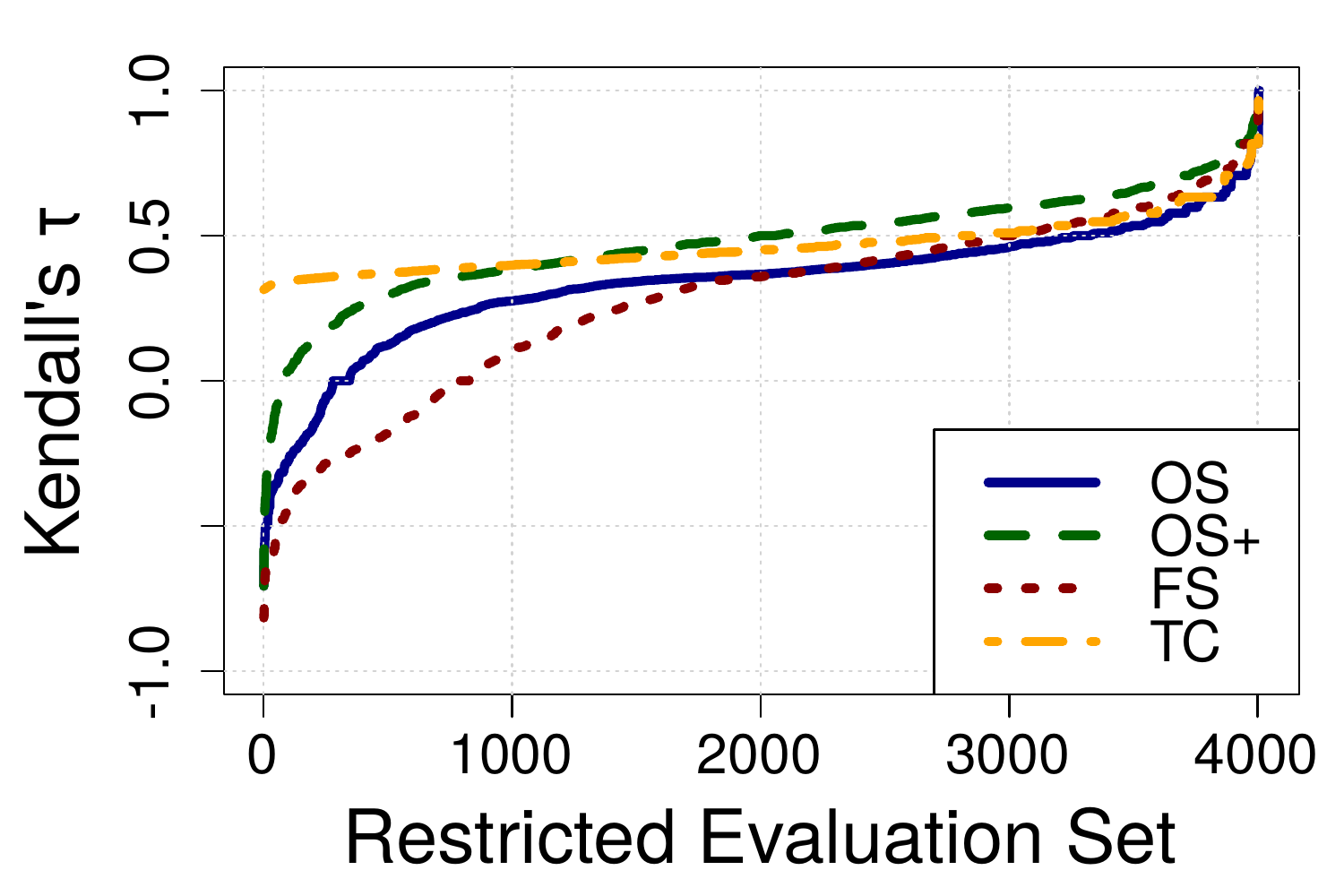}
\end{subfigure}%
\begin{subfigure}{.24\textwidth}
\includegraphics[width=\linewidth]{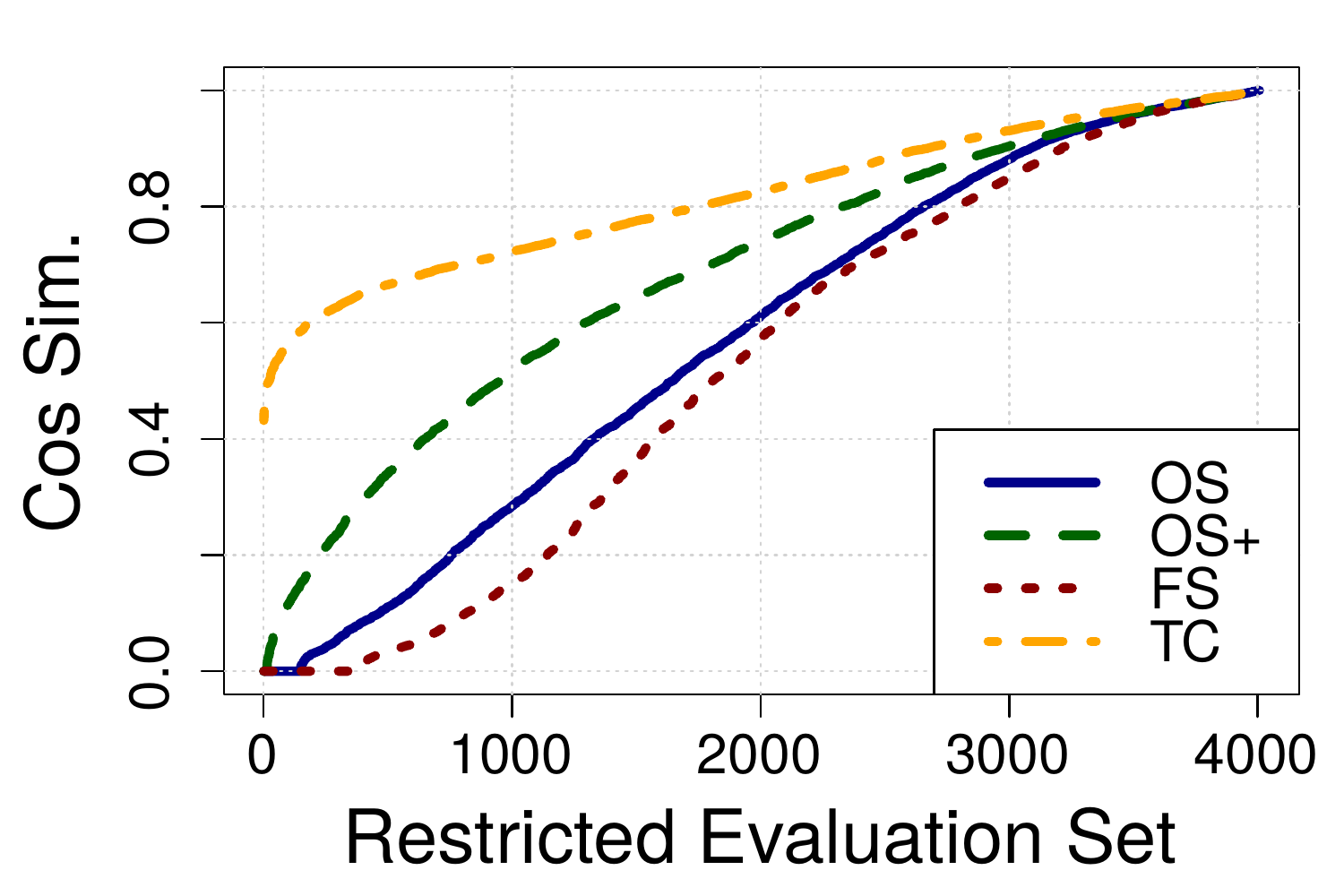}
\end{subfigure}
\begin{subfigure}{.24\textwidth}
\centering
\includegraphics[width=\linewidth]{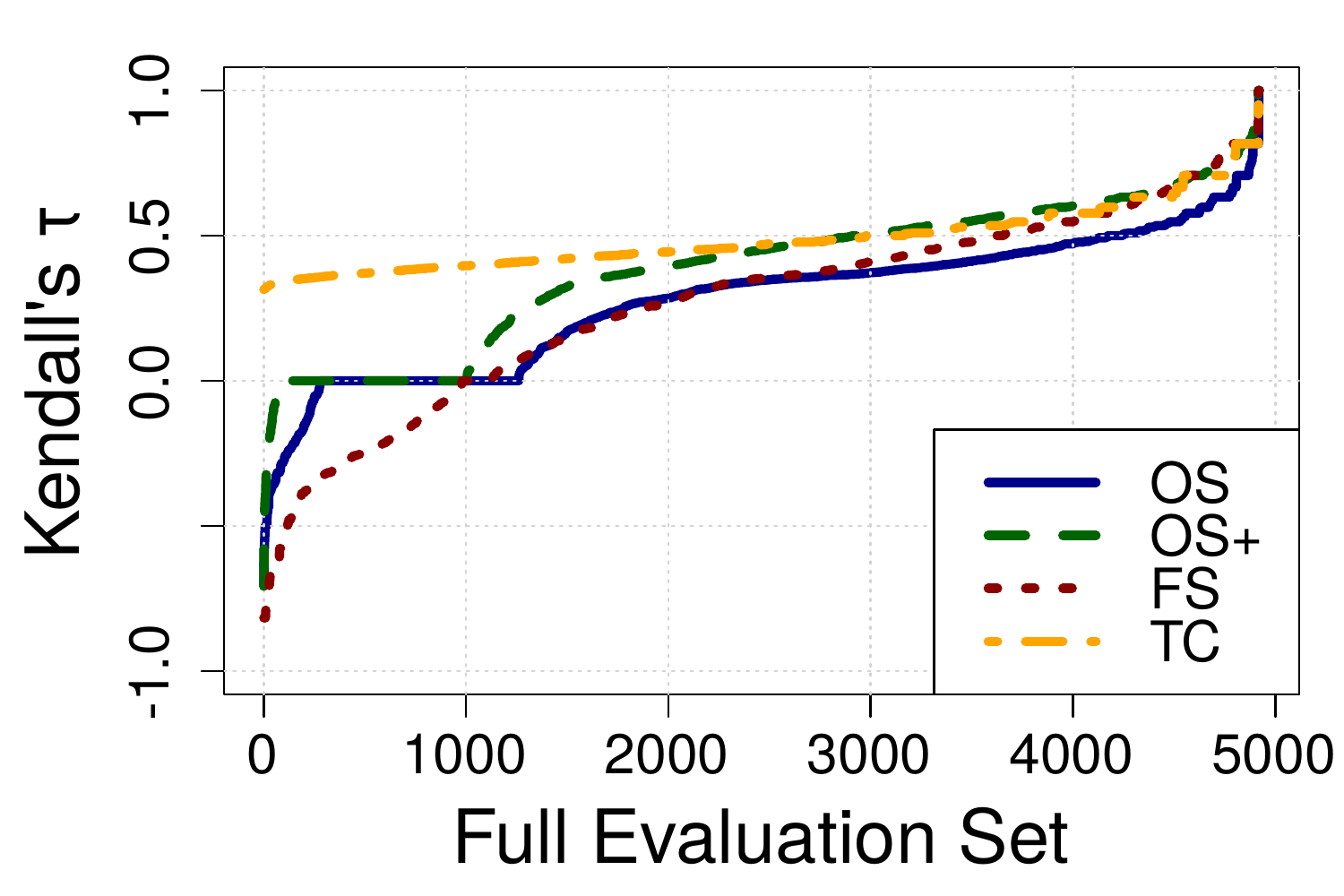}
\end{subfigure}%
\begin{subfigure}{.24\textwidth}
\includegraphics[width=\linewidth]{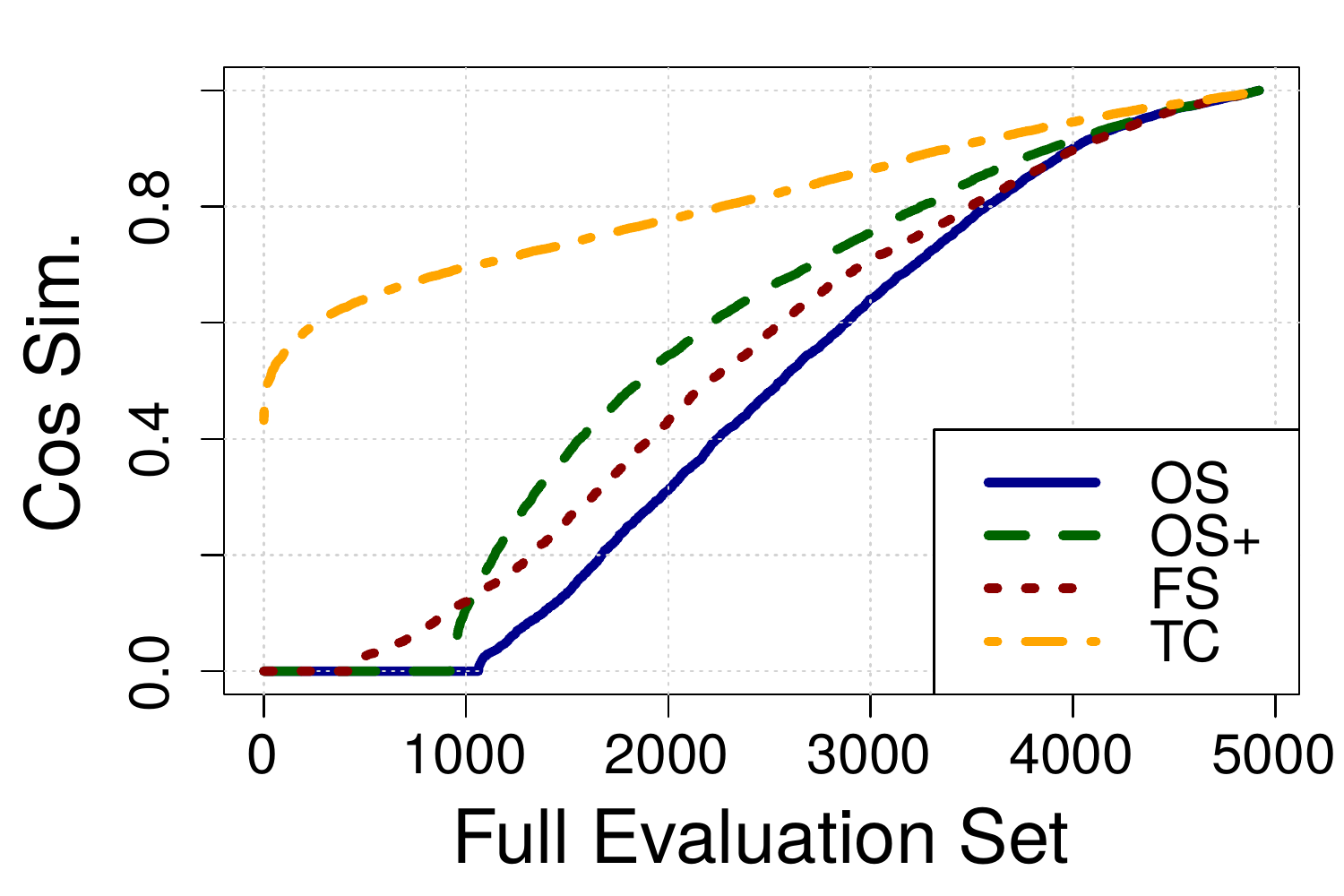}
\end{subfigure}
\setlength{\abovecaptionskip}{-0.0005pt}
\caption{Baselines evaluation results.}
\label{fig:eval}
\end{figure}


\begin{table}[t!]
    \centering
    \scalebox{0.75}{
    \begin{tabular}{cccccc}
         \hline
          & \textsc{Eval. Metric} & \textbf{OS} & \textbf{OS+}  & \textbf{FS} & \textbf{TC} \\ \hline
                      & Kendall's $\uptau$  AUC & 0.339 & \textbf{0.477} & 0.279 & 0.466 \\
         \textsc{R-}  & Kendall's $\uptau$ w-avg & 0.362 & \textbf{0.497} & 0.3 & 0.48 \\
         \textsc{Set} & Cos Sim AUC & 0.57 & \textbf{0.685} & 0.518 & 0.818 \\ 
                      & Cos Sim w-avg & 0.608 & \textbf{0.717} & 0.545 & 0.854 \\ \hline
         
                     & Kendall's $\uptau$ AUC & 0.269 & \textbf{0.379} & 0.253 & 0.491 \\
         \textsc{F-} & Kendall's $\uptau$ w-avg & 0.307 & \textbf{0.421} & 0.284 & 0.501 \\
         \textsc{Set} & Cos Sim AUC & 0.453 & \textbf{0.544} & 0.511 & 0.810 \\
                      & Cos Sim w-avg & 0.515 & \textbf{0.607} & 0.539 & 0.849 \\ \hline
    \end{tabular}
    }
    \caption{Aggregated evaluation results.}
    \label{tab:results}
\end{table}

The results (Figure~\ref{fig:eval} \& Table~\ref{tab:results}) show that OS+ performs best out of all the models, even taking into account sentences with LUs not in FN for which OS+ cannot disambiguate. FS performs the worst out of all models on \textsc{R-Set}, because it cannot find newly added frames from the latest FN release, but improves on the \textsc{F-Set} (FS can find candidate frames for LUs not in FN). The scores on the \textsc{F-Set} were lower for all baselines, suggesting that sentences with LUs not in FN are more difficult to classify -- this could be because FN is missing frames that can express the full meaning of these LUs. TC has a good performance, but is far from being unbeatable -- when measuring Kendall's $\uptau$ over the \textsc{R-Set}, OS+ performs better than TC.



\section{Conclusions}

We described a FrameNet frame disambiguation resource of 5,042 sentence-word pairs, and 1,000 LUs that are new to FN -- the largest corpus of this type outside of FN. Since the submission of this paper, the corpus has grown to over 9,000 sentence-word pairs. We also provide confidence scores for each candidate frame that are based on inter-worker disagreement. We made a case for this kind of disagreement reflecting genuine cases of ambiguity in FrameNet frames, caused by: child-parent relations between frames, frames with overlapping definitions, or compositions of frames making up the meaning of a word. The evaluation method we proposed uses the scores for multiple frames, and is thus able to differentiate between frames that still apply to the word, but with less clarity, and frames that capture the wrong meaning. Our goal was to build a resource that recognizes different levels of ambiguity in the expression of the frames in the text, and allows a more fair evaluation of performance of frame disambiguation systems.




\section*{Acknowledgments}

We would like to thank Luigi Asprino, Valentina Presutti and Aldo Gangemi for their assistance with using the Framester corpus, as well as their advice in better understanding the task of frame disambiguation. We would also like to thank the anonymous crowd workers for their contributions to our crowdsourcing tasks.

\bibliography{naaclhlt2019}

\begin{thebibliography}{18}
\expandafter\ifx\csname natexlab\endcsname\relax\def\natexlab#1{#1}\fi

\bibitem[{Aroyo and Welty(2014)}]{aroyo2014threesides}
Lora Aroyo and Chris Welty. 2014.
\newblock \href {https://doi.org/10.15346/hc.v1i1.3} {{The Three Sides of
  CrowdTruth}}.
\newblock \emph{Journal of Human Computation}, 1:31--34.

\bibitem[{Baker(2012)}]{baker2012framenet}
Collin~F Baker. 2012.
\newblock {FrameNet, current collaborations and future goals}.
\newblock \emph{Language Resources and Evaluation}, 46(2):269--286.

\bibitem[{Baker et~al.(1998)Baker, Fillmore, and Lowe}]{baker1998berkeley}
Collin~F Baker, Charles~J Fillmore, and John~B Lowe. 1998.
\newblock {The Berkeley FrameNet project}.
\newblock In \emph{Proceedings of the 17th international conference on
  Computational linguistics-Volume 1}, pages 86--90. Association for
  Computational Linguistics.

\bibitem[{Chang et~al.(2015)Chang, Paritosh, Huynh, and
  Baker}]{chang2015scaling}
Nancy Chang, Praveen Paritosh, David Huynh, and Collin Baker. 2015.
\newblock Scaling semantic frame annotation.
\newblock In \emph{Proceedings of The 9th Linguistic Annotation Workshop},
  pages 1--10.

\bibitem[{Dumitrache et~al.(2018{\natexlab{a}})Dumitrache, Aroyo, and
  Welty}]{DBLP:conf/hcomp/DumitracheAW18}
Anca Dumitrache, Lora Aroyo, and Chris Welty. 2018{\natexlab{a}}.
\newblock \href {https://aaai.org/ocs/index.php/HCOMP/HCOMP18/paper/view/17923}
  {Capturing ambiguity in crowdsourcing frame disambiguation}.
\newblock In \emph{{Proceedings of the Sixth {AAAI} Conference on Human
  Computation and Crowdsourcing, {HCOMP} 2018, Z{\"{u}}rich, Switzerland, July
  5-8, 2018.}}, pages 12--20. {AAAI} Press.

\bibitem[{Dumitrache et~al.(2018{\natexlab{b}})Dumitrache, Inel, Aroyo,
  Timmermans, and Welty}]{dumitrache2018crowdtruth}
Anca Dumitrache, Oana Inel, Lora Aroyo, Benjamin Timmermans, and Chris Welty.
  2018{\natexlab{b}}.
\newblock {CrowdTruth 2.0: Quality Metrics for Crowdsourcing with
  Disagreement}.
\newblock \emph{arXiv preprint arXiv:1808.06080}.

\bibitem[{FitzGerald et~al.(2015)FitzGerald, T{\"a}ckstr{\"o}m, Ganchev, and
  Das}]{fitzgerald2015semantic}
Nicholas FitzGerald, Oscar T{\"a}ckstr{\"o}m, Kuzman Ganchev, and Dipanjan Das.
  2015.
\newblock Semantic role labeling with neural network factors.
\newblock In \emph{Proceedings of the 2015 Conference on Empirical Methods in
  Natural Language Processing}, pages 960--970.

\bibitem[{Fossati et~al.(2013)Fossati, Giuliano, and
  Tonelli}]{fossati2013outsourcing}
Marco Fossati, Claudio Giuliano, and Sara Tonelli. 2013.
\newblock {Outsourcing FrameNet to the crowd}.
\newblock In \emph{Proceedings of the 51st Annual Meeting of the Association
  for Computational Linguistics (Volume 2: Short Papers)}, volume~2, pages
  742--747.

\bibitem[{Gangemi et~al.(2016)Gangemi, Alam, Asprino, Presutti, and
  Recupero}]{gangemi2016framester}
Aldo Gangemi, Mehwish Alam, Luigi Asprino, Valentina Presutti, and
  Diego~Reforgiato Recupero. 2016.
\newblock Framester: a wide coverage linguistic linked data hub.
\newblock In \emph{European Knowledge Acquisition Workshop}, pages 239--254.
  Springer.

\bibitem[{Hermann et~al.(2014)Hermann, Das, Weston, and
  Ganchev}]{hermann2014semantic}
Karl~Moritz Hermann, Dipanjan Das, Jason Weston, and Kuzman Ganchev. 2014.
\newblock Semantic frame identification with distributed word representations.
\newblock In \emph{Proceedings of the 52nd Annual Meeting of the Association
  for Computational Linguistics (Volume 1: Long Papers)}, volume~1, pages
  1448--1458.

\bibitem[{Hong and Baker(2011)}]{Hong:2011:GCR:2018966.2018970}
Jisup Hong and Collin~F. Baker. 2011.
\newblock \href {http://dl.acm.org/citation.cfm?id=2018966.2018970} {{How good
  is the crowd at ``real'' WSD?}}
\newblock In \emph{Proceedings of the 5th Linguistic Annotation Workshop}, LAW
  V '11, pages 30--37. Association for Computational Linguistics.

\bibitem[{J~Fillmore(1982)}]{Fillmore:1982}
Charles J~Fillmore. 1982.
\newblock \href {https://doi.org/10.1016/B0-08-044854-2/00424-7} {\emph{Frame
  Semantics}}, volume~34, pages 111--138.

\bibitem[{Jurgens(2013)}]{jurgens2013embracing}
David Jurgens. 2013.
\newblock Embracing ambiguity: A comparison of annotation methodologies for
  crowdsourcing word sense labels.
\newblock In \emph{HLT-NAACL}, pages 556--562.

\bibitem[{Kendall(1938)}]{kendall1938new}
Maurice~G Kendall. 1938.
\newblock A new measure of rank correlation.
\newblock \emph{Biometrika}, 30(1/2):81--93.

\bibitem[{Miller(1995)}]{miller1995wordnet}
George~A Miller. 1995.
\newblock {WordNet: a lexical database for English}.
\newblock \emph{Communications of the ACM}, 38(11):39--41.

\bibitem[{Pavlick et~al.(2015)Pavlick, Wolfe, Rastogi, Callison-Burch, Dredze,
  and Van~Durme}]{pavlick2015framenet+}
Ellie Pavlick, Travis Wolfe, Pushpendre Rastogi, Chris Callison-Burch, Mark
  Dredze, and Benjamin Van~Durme. 2015.
\newblock {FrameNet+: Fast paraphrastic tripling of FrameNet}.
\newblock In \emph{Proceedings of the 53rd Annual Meeting of the Association
  for Computational Linguistics and the 7th International Joint Conference on
  Natural Language Processing (Volume 2: Short Papers)}, volume~2, pages
  408--413.

\bibitem[{Snow et~al.(2008)Snow, O'Connor, Jurafsky, and Ng}]{Snow:2008}
Rion Snow, Brendan O'Connor, Daniel Jurafsky, and Andrew~Y. Ng. 2008.
\newblock Cheap and fast---but is it good?: evaluating non-expert annotations
  for natural language tasks.
\newblock In \emph{Proceedings of the Conference on Empirical Methods in
  Natural Language Processing}, EMNLP '08, pages 254--263. Association for
  Computational Linguistics.

\bibitem[{Swayamdipta et~al.(2017)Swayamdipta, Thomson, Dyer, and
  Smith}]{swayamdipta:17}
Swabha Swayamdipta, Sam Thomson, Chris Dyer, and Noah~A. Smith. 2017.
\newblock {Frame-Semantic Parsing with Softmax-Margin Segmental RNNs and a
  Syntactic Scaffold}.
\newblock \emph{arXiv preprint arXiv:1706.09528}.

\end{thebibliography}
\bibliographystyle{acl_natbib}

\onecolumn

\appendix

\section{Ambiguity Examples in the Corpus}

\begin{table*}[thb!]
    \centering
    \scalebox{0.75}{
    \begin{tabular}{cp{12cm}cp{5cm}}
        \hline
        \# & \textsc{Sentence} & \textbf{$SQS$} & \textsc{Frames ($FSS$)}  \\ \hline
        1 & These Articles have historically shaped and \textbf{continue} to direct the ethos of the Communion. & 0.795 & \textit{activity ongoing} (0.862) \newline \textit{process continue} (0.86) \\
        2 & ``A Modest Proposal'' is \textbf{included} in many literature programs as an example of early modern western satire. & 0.771 & \textit{inclusion} (0.89) \newline \textit{cause to be included} (0.813) \\ 
        3 & The states often \textbf{failed} to meet these requests in full, leaving both Congress and the Continental Army chronically short of money. & 0.628 & \textit{endeavor failure} (0.826) \newline \textit{success or failure} (0.8) \\
        4 & This is a chart of trend of nominal gross domestic product of Angola at market prices \textbf{using} International Monetary Fund data. & 0.598 & \textit{using resource} (0.831) \newline \textit{using} (0.554) \newline \textit{tool purpose} (0.336) \\
        5 & The Asian tigers have now all \textbf{received} developed country status, having the highest GDP per capita in Asia. & 0.504 & \textit{receiving} (0.751) \newline \textit{getting} (0.556) \\
        6 & MasterCard has released Global Destination Cities Index 2013 with 10 of 20 are \textbf{dominated} by Asia and Pacific Region Cities. & 0.467 & \textit{dominate situation} (0.638) \newline \textit{dominate competitor} (0.579) \newline \textit{being in control} (0.327) \\ 
        \hline
    \end{tabular}
    }
    \caption{Ambiguity because of parent-child relation between frames.}
    \label{tab:sub_super}
\end{table*}

\begin{table*}[thb!]
    \centering
    \scalebox{0.75}{
    \begin{tabular}{cp{12cm}cp{5cm}}
        \hline
        \# & \textsc{Sentence} & \textbf{$SQS$} & \textsc{Frames ($FSS$)}  \\ \hline
        1 & Kournikova then \textbf{withdrew} from several events due to continuing problems with her left foot and did not return until Leipzig. & 0.725 & \textit{withdraw from participation} (0.955), \textit{removing} (0.61) \\
        2 & Some aikido organizations use belts to \textbf{distinguish} practitioners' grades. & 0.68 & \textit{differentiation} (0.867) \newline \textit{distinctiveness} (0.703) \\
        3 & Since then, it has focused on \textbf{improving} relationships with Western countries, cultivating links with other Portuguese-speaking countries, and asserting its own national interests in Central Africa. & 0.654 & \textit{improvement or decline} (0.787) \newline \textit{cause to make progress} (0.732) \\ 
        4 & To \textbf{emphasize} the validity of the Levites' claim to the offerings and tithes of the Israelites, Moses collected a rod from the leaders of each tribe in Israel and laid the twelve rods over night in the tent of meeting. & 0.65 & \textit{emphasizing} (0.764) \newline \textit{convey importance} (0.638) \\
        5 & He not only had enough food from his subjects to \textbf{maintain} his military, but the taxes collected from traders and merchants added to his coffers sufficiently to fund his continuous wars. & 0.453 & \textit{cause to continue} (0.7) \newline \textit{activity ongoing} (0.602) \\
        6 & He \textbf{spent} the later part of his life in the United States, living in Los Angeles from 1937 until his death. & 0.29 & \textit{taking time} (0.41) \newline \textit{expend resource} (0.365)  \\ 
        \hline
    \end{tabular}
    }
    \caption{Ambiguity because of overlapping frame definitions.}
    \label{tab:overlap}
\end{table*}

\begin{table*}[thb!]
    \centering
    \scalebox{0.75}{
    \begin{tabular}{cp{12cm}cp{5cm}}
        \hline
        \# & \textsc{Sentence} & \textbf{$SQS$} & \textsc{Frames ($FSS$)}  \\ \hline
        1 & These writings lack the mystical, philosophical elements of alchemy, but do contain the works of Bolus of Mendes (or Pseudo-Democritus), which \textbf{aligned} these recipes with theoretical knowledge of astrology and the classical elements. & 0.284 & \textit{arranging} (0.474) \newline \textit{adjusting} (0.4) \newline \textit{assessing} (0.298) \newline \textit{compatibility} (0.254) \newline \textit{undergo change} (0.169) \\
        2 & However, commercial application of this fact has challenges in \textbf{circumventing} the passivating oxide layer, which inhibits the reaction, and in storing the energy required to regenerate the aluminium metal. & 0.239 & \textit{dodging} (0.477) \newline \textit{compliance} (0.248) \newline \textit{surpassing} (0.204) \newline no frame (0.148) \\ 
        3 & This had the effect of \textbf{inculcating} the principle of ``Lex orandi, lex credendi'' (Latin loosely translated as 'the law of praying [is] the law of believing') as the foundation of Anglican identity and confession. & 0.201 & \textit{education teaching} (0.384) \newline \textit{communication} (0.35) \newline no frame (0.153) \\ 
        4 & Legal segregation ended in the states in 1964, but Jim Crow customs often continued until specifically \textbf{challenged} in court. & 0.172 & \textit{difficulty} (0.372) \newline \textit{competition} (0.283) \newline \textit{taking sides} (0.257) \newline \textit{communication} (0.154) \\ 
        5 & When Washington's army arrived outside Yorktown, Cornwallis prematurely abandoned his outer position, \textbf{hastening} his subsequent defeat. & 0.134 & \textit{speed description} (0.39) \newline \textit{assistance} (0.209) \newline \textit{self motion} (0.165) \newline \textit{travel} (0.16) \newline \textit{causation} (0.124) \\
        \hline
    \end{tabular}
    }
    \caption{Ambiguity because the meaning of the word is expressed by a composition of frames.}
    \label{tab:composition}
\end{table*}

\end{document}